%% file: _main.tex
\input{_constants}
\arxiv % \review OR \arxiv OR \cameraready

\pdfoutput=1
\documentclass[10pt,twocolumn,letterpaper]{article}
\input{cvpr_header}
\begin{document}
%% TITLE
\title{\paperTitle}
\author{\authorBlock}
\maketitle

\newcommand{\MN}[1]{{\textcolor{red}{[\textbf{MN:} #1]}}}

\input{00_abstract}
\input{01_intro}
\input{02_related}

\input{03_method}

\input{10_conclusion}

{\small
\bibliographystyle{ieeenat_fullname}
\bibliography{11_references}
}

\ifarxiv \clearpage \appendix \input{12_appendix} \fi

\end{document}

%% file: _constants.tex
\def\paperID{3793}
\def\confName{CVPR}
\def\confYear{2024}

\def\paperTitle{DNS SLAM: Dense Neural Semantic-Informed SLAM}

\def\authorBlock{
    Kunyi Li$^1$\qquad     %\thanks{Equal contribution} \qquad
    Michael Niemeyer$^2$\qquad
    Nassir Navab$^{1, 3}$\qquad
    Federico Tombari$^{1, 2}$\qquad\\
    \textsuperscript{1}Technical University of Munich\qquad
    \textsuperscript{2}Google\qquad
    \textsuperscript{3}Johns Hopkins University
}

\newif\ifreview 
\newif\ifarxiv \newcommand{\arxiv}{\arxivtrue}
\newif\ifcamera 
\newif\ifrebuttal 

%% file: cvpr_header.tex
\ifreview \usepackage[review]{cvpr} \fi
\ifarxiv \usepackage[pagenumbers]{cvpr} \fi
\ifrebuttal \usepackage[rebuttal]{cvpr} \fi
\ifcamera \usepackage{cvpr} \fi

\input{_macros}  % Add packages to _macros.tex

\usepackage{xr-hyper}

\makeatletter
\newcommand*{\addFileDependency}[1]{
  \typeout{(#1)}
  \@addtofilelist{#1}
  \IfFileExists{#1}{}{\typeout{No file #1.}}
}

\makeatother

\definecolor{cvprblue}{rgb}{0.21,0.49,0.74}
\usepackage[pagebackref,breaklinks,colorlinks,citecolor=cvprblue]{hyperref}
\usepackage[capitalize]{cleveref}
\crefname{section}{Sec.}{Secs.}
\crefname{table}{Table}{Tables}
\crefname{figure}{Fig.}{Figs.}

%% file: _macros.tex
\usepackage{graphicx}	
\usepackage{amsmath}	
\usepackage{amssymb}	
\usepackage{booktabs}
\usepackage{times}
\usepackage{microtype}
\usepackage{epsfig}
\usepackage[table,xcdraw,dvipsnames]{xcolor}
\usepackage{caption}
\usepackage{float}
\usepackage{placeins}
\usepackage{color, colortbl}
\usepackage{stfloats}
\usepackage{enumitem}
\usepackage{tabularx}
\usepackage{xstring}
\usepackage{multirow}
\usepackage{xspace}
\usepackage{url}
\usepackage{subcaption}
\usepackage{xcolor}
\usepackage[hang,flushmargin]{footmisc}

\ifcamera \usepackage[accsupp]{axessibility} \fi

% ...

\ifarxiv  \fi

\newcommand{\R}[1]{{%
    \textbf{%
        \ifstrequal{#1}{1}{\textcolor{red}{R#1}}{%
        \ifstrequal{#1}{2}{\textcolor{blue}{R#1}}{%
        \ifstrequal{#1}{3}{\textcolor{magenta}{R#1}}{%
        \ifstrequal{#1}{4}{\textcolor{teal}{R#1}}{%
                           \textcolor{cyan}{R#1}%
        }}}}%
    }%
}}

%% file: 00_abstract.tex
\begin{abstract}
% Abstract goes here.
In recent years, coordinate-based neural implicit representations have shown promising results for the task of Simultaneous Localization and Mapping (SLAM). While achieving impressive performance on small synthetic scenes, these methods often suffer from oversmoothed reconstructions, especially for complex real-world scenes. 
In this work, we introduce DNS SLAM, a novel neural RGB-D semantic SLAM approach featuring a hybrid representation. 
Relying only on 2D semantic priors, we propose the first semantic neural SLAM method that trains class-wise scene representations while providing stable camera tracking at the same time. Our method integrates multi-view geometry constraints with image-based feature extraction to improve appearance details and to output color, density, and semantic class information, enabling many downstream applications.
To further enable real-time tracking, we introduce a lightweight coarse scene representation which is trained in a self-supervised manner in latent space. 
Our experimental results achieve state-of-the-art performance on both synthetic data and real-world data tracking while maintaining a commendable operational speed on off-the-shelf hardware. Further, our method outputs class-wise decomposed reconstructions with better texture capturing appearance and geometric details.
\end{abstract}

%% file: 01_intro.tex
\label{sec:intro}
\section{Introduction}

% To insert a figure: \input{figs/template}
% Or table: \input{tables/template}

\begin{figure}
\includegraphics[width=8.5cm]{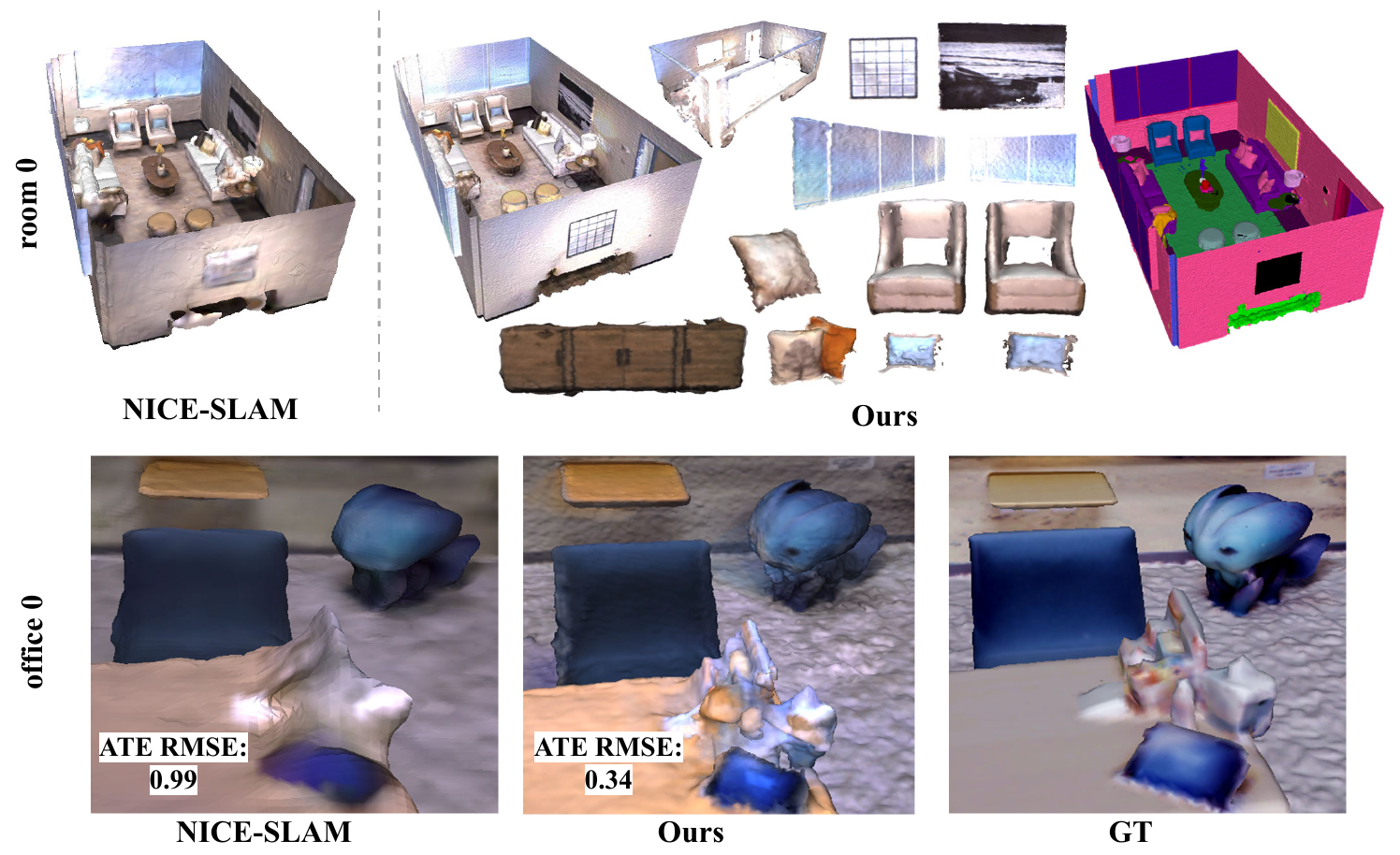}
\centering
\caption{\textbf{Class-Wise Reconstruction using DNS SLAM.} In contrast to previous neural SLAM systems, our method leads to semantically-decomposed 3D reconstructions of the scene (top) required for many downstream applications. Further, by using image-based features, we improve camera tracking and our reconstructions contain more geometric and appearance details (bottom).}
%Our method achieve accurate tracking and class-wise reconstruction. Details are well preserved even for tiny objects.
\label{fig:start}
\end{figure}

Dense Visual Simultaneous Localization and Mapping (SLAM) is a fundamental problem in the field of computer vision and plays a crucial role in various applications such as autonomous driving, indoor robotics, mixed reality, and more. Its primary goal is to create a 3D map of an unknown environment while simultaneously estimating camera poses. Traditional SLAM systems \cite{davison2007monoslam, mur2015orb, klein2007parallel, izadi2011kinectfusion} rely on multi-view geometry and focus more on camera pose accuracy but are prone to trajectory drifting. Recent learning-based dense visual SLAM methods \cite{davison2007monoslam, weder2020routedfusion, weder2021neuralfusion, huang2021di,teed2021droid, sucar2021imap, zhu2022nice, johari2023eslam, wang2023co, sandstrom2023point, Hu2023LNI-ADFP} have aimed instead to generate meaningful global 3D maps, albeit with limited reconstruction accuracy and demanding extensive training for accurate pose estimation.

Inspired by the advancements in neural field research, neural implicit SLAM methods like iMAP \cite{sucar2021imap} and NICE-SLAM \cite{zhu2022nice} have recently emerged. These methods estimate poses from a scene representation, prioritizing world-space geometry over image-space geometry, showing promising results on synthetic indoor datasets. They leverage the inherent smoothness and coherence encoded in Multilayer Perceptron (MLP) weights, making them suitable for sequential tracking and mapping tasks. However, these methods tend to oversmooth details in the reconstruction, which causes additional tracking errors.
To address this challenge, Co-SLAM \cite{wang2023co} uses parametric embeddings, while Point-SLAM \cite{sandstrom2023point} combines point cloud data with neural implicit representations to enhance reconstruction details. However, we find that their remarkable performance heavily relies on accurate pixel-perfect depth supervision. But in the real-world the depth is usually noisy, and we observe that these methods degrate drastically in these scenarios.

For robotics and interactive real-world vision applications, a semantic model that represents scene classes separately is required. vMAP \cite{kong2023vmap} excels at accurate and complete scene reconstruction by modeling objects separately, but focuses only on reconstruction and does not perform localization.

To bridge the gap and fully leverage multi-view geometry and semantic information, we introduce Dense Neural Semantic-Informed (DNS) SLAM. Our method utilizes class-wise scene representations which can build constraints between each class and the current camera pose. Further, 2D features from reference images provide multi-view geometry constraints for better camera pose estimation. In summary, our main \textbf{contributions} include:
\begin{itemize}
\item Leveraging 2D semantic priors, we investigate a semantic-informed multi-class scene representation, yielding an efficient, comprehensive, and semantically decomposed geometry representation.
\item Utilizing multi-view geometry, we extract image features by back-projecting points into reference frames, establishing constraints on relative camera poses and enhancing the appearance details.
\item To speed up tracking, we introduce a lightweight coarse scene representation which is trained with a novel self-supervision strategy, utilizing the multi-class representation as pseudo ground-truth.
\item To achieve accurate and smooth reconstructions, we approximate occupancy probabilities with Gaussian distributions as the ground-truth for additional geometry supervision.
\end{itemize}
Compared with state-of-the-art (SOTA) methods, our method achieves better performance on camera pose estimation on both synthetic and real-world datasets, improving ATE RMSE over 10\% on average. 

%% file: 02_related.tex
\section{Related Work}
\label{sec:related}

\textbf{Scene Representations.}
Since the introduction of neural fields in the context of 3D reconstruction~\cite{MeschederONNG19,ChenZ19,ParkFSNL19},
they have been applied to many 3D computer vision tasks, including novel view synthesis \cite{mildenhall2022nerf}, semantic scene reconstruction \cite{zhi2021place}, and camera pose estimation \cite{bian2023nope}. 
Approaches such as Neural Radiance Fields (NeRF)~\cite{mildenhall2021nerf} have achieved remarkable scene reconstruction results through differentiable rendering. However, it's worth noting that coordinate-based encoding methods \cite{zhu2022nice, izadi2011kinectfusion, dai2017bundlefusion, weder2020routedfusion, weder2021neuralfusion, whelan2015elasticfusion, yu2022monosdf, guo2022neural, azinovic2022neural} come with the drawback of extended training times.
To mitigate this, recent approaches \cite{muller2022instant, niessner2013real, kahler2015hierarchical} employ parametric encoding techniques expediting the training process. This enables neural implicit methods for real-time SLAM applications. While some works focus on improving the efficiency, Pixel-NeRF \cite{yu2021pixelnerf} preserves the spatial alignment between images and 3D representation, which is also an important concept in SLAM, by extracting 2D features from reference images and operating in view-space.
Semantic-NeRF \cite{zhi2021place} provides precise geometry and semantic outcomes even with minimal semantic guidance. This demonstrates that with solely 2D semantic supervision, neural implicit methods can generate compelling 3D semantic scene reconstruction. For real-world SLAM applications, a semantic model is undoubtedly more meaningful. Our work partly builds upon the foundations of both Pixel-NeRF and Semantic-NeRF, but takes a significant step forward by simultaneously optimizing the camera poses while they assuming pose data is given.

\noindent \textbf{Dense Visual SLAM.} 
Modern SLAM methods follow the overall architecture introduced by MonoSLAM \cite{davison2007monoslam}, PTAM \cite{klein2007parallel} and ORB SLAM \cite{mur2015orb}, decomposing the task into mapping and tracking. Based on the working space of map reconstruction, SLAM methods can be generally divided into two categories: image view-centric space and world view-centric space. The former one aims at maintaining multi-view geometry among keyframes in the dense setting, and the later one anchors the 3D geometry representation in a uniform world coordinate.
DTAM \cite{newcombe2011dtam} is an early example of image view-centric methods and has been adapted in many recent learning-based SLAM system \cite{teed2018deepv2d, zhi2019scenecode, bloesch2018codeslam, sucar2020nodeslam}. DeepTAM \cite{zhou2018deeptam} combines the cost volume and keyframe image to update the depth prediction. D3VO \cite{yang2020d3vo} models the photometric uncertainties of pixels on the input images, which improves the depth estimation accuracy. And Droid SLAM \cite{teed2021droid} uses optical flow to define geometrical residuals.
World view-centric methods store global map in surfels \cite{schops2019bad, whelan2015elasticfusion}, octrees \cite{vespa2018efficient} or grids like voxel grid \cite{izadi2011kinectfusion, dai2017bundlefusion, weder2020routedfusion, weder2021neuralfusion, whelan2015elasticfusion} and hash grid \cite{kahler2015hierarchical, niessner2013real}. 
KinectFusion \cite{izadi2011kinectfusion} performs a frame-to-model camera tracking strategy and updates the scene geometry via Truncated Signed Distance Function (TSDF) fusion. RoutedFusion \cite{weder2020routedfusion} outputs the TSDF update of the volumetric grid. NeuralFusion \cite{weder2021neuralfusion} extends this concept by learning the scene representation implicitly. 
In our method, the benefits from both image view and world are adopted, using a hybrid scene representation.

\noindent \textbf{Neural Implicit SLAM.}
More recently, there has been a popularity in neural implicit representations for dense visual SLAM. iMAP \cite{sucar2021imap}, as the first neural implicit SLAM method, employs an MLP-based representation to conduct joint tracking and mapping in quasi-real time, but the reconstruction and camera pose result are far worse than traditional methods.
To address computational overhead and scalability, NICE-SLAM \cite{zhu2022nice} introduces a multi-level feature grid as scene representation. Nevertheless, the feature grid's local update approach limits its hole-filling capabilities. 
On the other hand, Co-SLAM \cite{wang2023co} attains real-time performance by combining coordinate and sparse parametric encodings for scene representation, employing dense global bundle adjustment using rays sampled from all keyframes.
E-SLAM \cite{johari2023eslam} suggests the use of multi-scale axis-aligned feature planes to prevent the model size from growing cubically concerning the scene's size.
Point-SLAM \cite{sandstrom2023point} introduces a neural point cloud that iteratively grows in a data-driven manner during scene exploration for both mapping and tracking. ADFP \cite{Hu2023LNI-ADFP} proposes an attentive depth fusion prior and uses either currently learned geometry or the one from depth fusion in volume rendering.

%% file: 03_method.tex
\section{Method}
\label{sec:method}

\begin{figure*}[t]
\includegraphics[width=17cm]{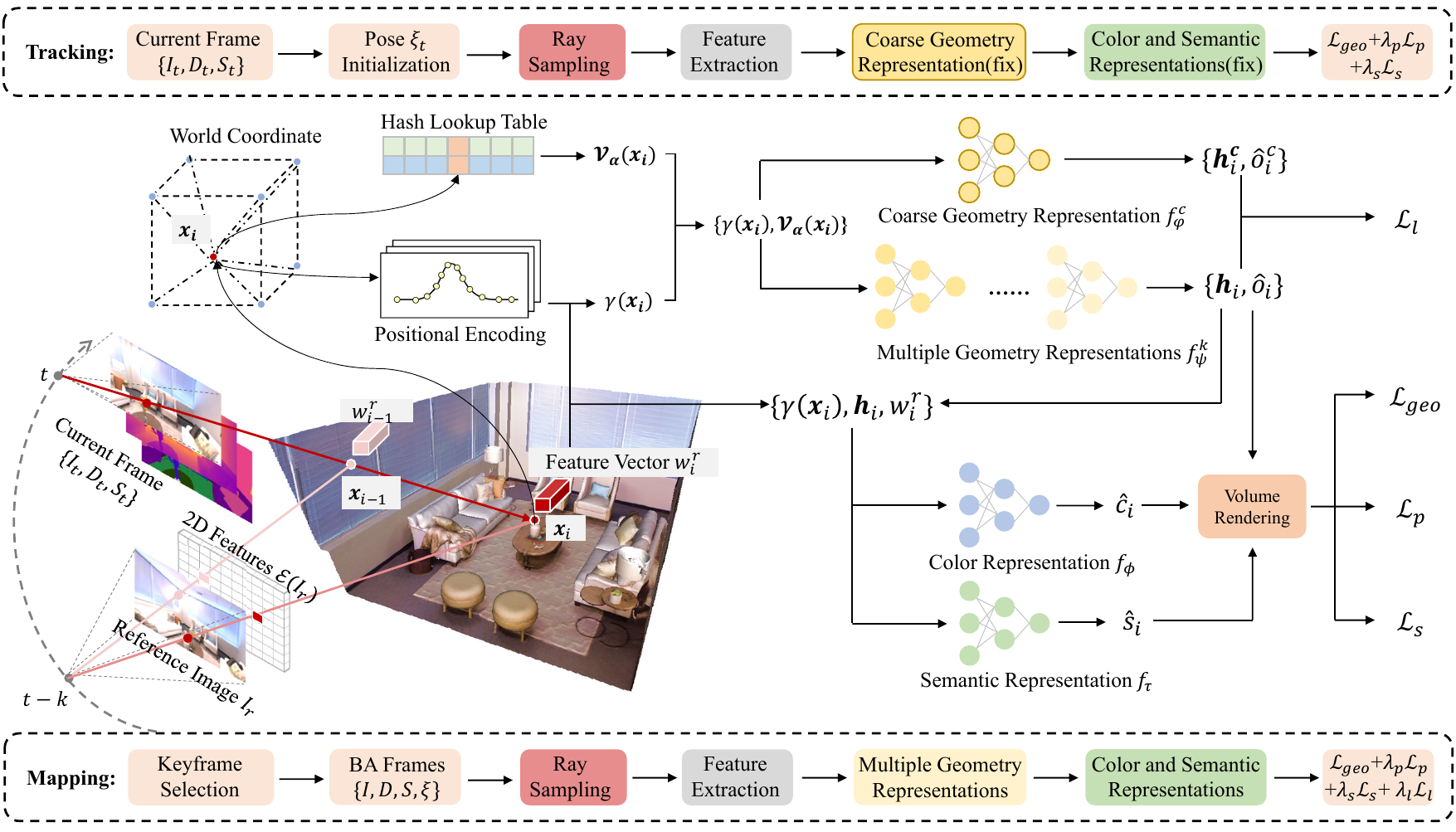}
\centering
\caption{\textbf{Overview.} 1) Scene Representation: For each point along the ray, we first query a 3D feature from our hash-based feature grid and a 2D feature from reference image(s). Next, we use multiple geometry together with color and semantic representations to render occupancy, color, semantic logits to 2D. 2) Tracking: We optimize per-frame camera poses by utilizing a coarse geometry representation which in turn is trained in a self-supervised manner using our fine decomposed representation. 3) Mapping: In global and local bundle adjustment, we jointly optimize the scene representation and camera poses.}
\label{fig:overview}
\end{figure*}

\subsection{Scene Representation} 
\label{subsec:overview}
\noindent \textbf{Notation}:
Let $\xi_t^{c2w}=[R_t^{c2w}|\mathbf{t}_t^{c2w}]\in SE(3)$ be the camera-to-world transform of frame $t$, where $R_t^{c2w}\in SO(3)$ and $\mathbf{t}_i^{c2w}\in\mathbb{R}^3$ are the rotation and translation, respectively. Denote camera intrinsic matrix as $K\in\mathbb{R}^{3\times3}$. 
For the rest of the manuscript 
%unless otherwise stated, 
we set $\xi=\xi^{c2w}$ to avoid cluttered notation.
Camera projection and back-projection operator is denoted as $\pi^{-1}$ and $\pi$ , respectively, where $\pi$ maps a 3D point $\mathbf{x}^c$ in the camera coordinate system into the image coordinate system and $\pi^{-1}$ maps a pixel $\mathbf{u}$ in image coordinate system with depth $d(\mathbf{u})$ into the camera coordinate system:
\begin{equation}
    \pi(\mathbf{x}^c)\cong K\mathbf{x}^c, \pi^{-1}(\mathbf{u}, d(\mathbf{u}))\cong d(\mathbf{u})K^{-1}{\mathbf{u}}
\end{equation}
Note that we drop explicit homogeneous notation for clarity.

\noindent \textbf{Neural Rendering}:
The input of our SLAM system is a RGB-D and semantic label stream $\{I_t\}^T_{t=1}, \{D_t\}^T_{t=1}, \{S_t\}^T_{t=1}$ with known intrinsics $K$. Our goal is to estimate camera poses $\{\hat{\xi}_t\}^T_{t=1}$ and train a scene representation $\theta$. 
Each pixel in image space $\mathbf{u}\in\mathbb{R}^2$ determines a ray in world space with origin $\mathbf{o}=\mathbf{\hat{t}}_t$ and ray direction $\mathbf{d}={\hat{R}_t}K^{-1}\mathbf{u}$.
For rendering, we randomly sample $M$ points $\mathbf{x}_i=\mathbf{o}+d_i\mathbf{d}, i\in\{1, ..., M\}$ along the ray where $d_i$ denotes depth value of sampled point $\mathbf{x}_i$. 
%Same as in Co-SLAM\cite{wang2023co},
We use One-blob \cite{muller2019neural} encoding $\gamma(\mathbf{x}_i)$ similar to~\cite{wang2023co} and a multi-resolution hash-based feature grid $\mathcal{V}_\alpha=\{\mathcal{V}_\alpha^l\}^L_{l=1}$ \cite{muller2022instant} as geometry representation. Our method first maps $\mathbf{x}_i$ into occupancy value $\hat{o}_i$ and latent vector $\mathbf{h}_i$ with geometry representation:
\begin{equation}
\label{eq:geo nerf}
    f_\psi(\gamma(\mathbf{x}_i), \mathcal{V}_\alpha(\mathbf{x}_i)) \mapsto (\mathbf{h}_i, \hat{o}_i).
\end{equation}
We use multiple geometry representations $f_\psi^k$ for each class in the scene instead of single one, which will be explained in Sec. \ref{subsec:multi-class}. Then, the color representation predicts color value $\hat{\mathbf{c}}_i$ as: 
\begin{equation}
\label{eq:color nerf}
    f_\phi(\gamma(\mathbf{x}_i), \mathbf{h}_i) \mapsto \hat{\mathbf{c}}_i.
\end{equation}
Besides geometry and appearance, we further use a semantic representation to predict semantic logit value $\hat{\mathbf{s}}_i$ as:
\begin{equation}
\label{eq:semantic nerf}
    f_\tau(\gamma(\mathbf{x}_i), \mathbf{h}_i) \mapsto \hat{\mathbf{s}}_i.
\end{equation}
We set all learnable parameters of our scene representation as $\theta=\{\alpha, \psi, \phi, \tau\}$.
%Following UNISURF \cite{oechsle2021unisurf}, 
We render depth $\hat{d}$, color $\hat{\mathbf{c}}$ and semantic logit value $\hat{\mathbf{s}}$ by integrating the predicted values along the sample rays~\cite{mildenhall2021nerf,oechsle2021unisurf}:
\begin{equation}
\label{eq:rendering}
    \mathbf{\hat{g}}=\mathbf{\hat{g}}(\mathbf{u}; \hat{\xi}_t, \theta)=\sum_{i=1}^{M}w_i\hat{\mathbf{g}}_i,
\end{equation}
where $\mathbf{\hat g}\in\{\mathbf{\hat c}, \hat d, \mathbf{\hat s}\}$. The weight $w_i$ represents the discretized probability that the ray terminates at point $\mathbf{x}_i$:
\begin{equation}
    w_i=\hat{o}_i\prod_{j=1}^{i-1}(1-\hat{o}_i).
\end{equation}
Like traditional SLAM methods \cite{mur2015orb}, our method can be split into two processes: tracking and mapping. Tracking process first estimates camera pose of each frame  while keeping the scene representations fixed. Mapping process jointly optimizes both the scene representations and camera poses. More details will be explained in the following sections. Fig.\ \ref{fig:overview} shows an overview of our method.

\subsection{Multi-Class Consistency}
\label{subsec:multi-class}
Instead of using a single shared MLP similar to previous neural SLAM approaches, decomposing the prediction into smaller class-specific MLPs can lead to to a significant performance boost and speedup~\cite{ReiserPL021,kundu2022panoptic, kong2023vmap}. While obtaining 3D semantic priors is difficult in most cases, 2D semantic segmentation \cite{long2015fully, ronneberger2015u, zhou2022detecting} is a well-studied task. As a result, in our method, we choose to utilize 2D semantic maps as they are often directly provided in datasets or alternatively can also be obtained from off-the-shelf methods. 
While we focus on class-level semantic maps in this work, other decompositions such as instance-level maps could be further explored in the future.
%The 2D semantic map can be either class level or object/instance level. Since class-specific MLPs represent each class separately, a camera pose consistency can be built among each class. 

\noindent \textbf{Multi-Class Rendering:} For each pixel $\mathbf{u}$ in the image, first the 2D semantic map is queried to obtain its corresponding class id $k=\mathbf{s}(\mathbf{u})$, where $\mathbf{s}(\mathbf{u})$ denotes ground-truth semantic label at pixel  $\mathbf{u}$. In this case, scene representation Eq. \ref{eq:geo nerf} can be rewritten as:
\begin{equation}
\label{eq:fine nerf}
    f_\psi^k(\gamma(\mathbf{x}_i), \mathcal{V}_\alpha(\mathbf{x}_i)) \mapsto (\mathbf{h}_i, \hat{o}_i),
\end{equation}
and density can be rendered as:
\begin{equation}
    \hat{d}=\hat{d}(\mathbf{u}; \hat{\xi}, \psi^k)=\sum_{i=1}^{M}w_i\hat{o}_i,
\end{equation}
where $k=1, ..., K$, denotes different class. Therefore, the geometry loss function can be expressed as:
\begin{equation}
\begin{matrix}
\label{eq:geo loss}
    \min & \mathcal{L}_{geo}=\sum_{\mathbf{u}\in|I|}\left \| d(\mathbf{u})- \hat{d}(\mathbf{u}; \hat{\xi}, \theta)\right \|_1 %, \\
    %.t. & \ \hat{\xi}_t^1=...=\hat{\xi}_t^k,
\end{matrix}
\end{equation}
where $d(\mathbf{u})$ denotes ground-truth depth value at pixel $\mathbf{u}$, and $|I|$ represents a set of pixels sampled from image $I$. We use the $L_1$ norm as our geometry loss $\mathcal{L}_{geo}$.

\subsection{Multi-View Consistency}
\label{subsec:multi-view}
Previous neural implicit SLAM methods 
%use color images as the only supervision for rendered color value, 
directly predict RGB colors from latent representations and they tend to lead to oversmoothed reconstructions. As shown in image-based rendering approaches such as Pixel-NeRF \cite{yu2021pixelnerf}, spatial image features aligned to each pixel as an input allows the model to learn not only more texture information, but also scene priors. Therefore, we use features from 2D images as a conditional input in our model.

\noindent \textbf{Image-Based Feature Pooling:} Specifically, for every target frame $I$, we choose a reference frame $I_r$ and extract 2D features $\mathcal{E}(I_r)$ using off-the-shelf method (ResNet-18~\cite{he2016deep} in our case). Next, we cast rays for the target frame $I$ to get sets of 3D points $\{\mathbf{x}_i\}$, and each 3D point $\mathbf{x}_i={\hat{\xi}_t}\pi^{-1}(\mathbf{u}, d_i(\mathbf{u}))$ is back-projected to the reference frame coordinate to retrieve the corresponding image feature vectors $\mathbf{w}^r_i$:
\begin{equation}
    \mathbf{w}^r_i=\mathcal{E}(I_r(\pi(({\hat{\xi}_r})^{-1}{\hat{\xi}_t}\pi^{-1}(\mathbf{u}, d_i(\mathbf{u}))))).
\end{equation}
Instead of directly passing these features to our scene representation, we encode each feature with the corresponding reference view $(\mathbf{o}_r, \mathbf{d}_r)$ to obtain intermediate feature vectors:
\begin{equation}
    f_\sigma(\gamma(\mathbf{o}_r, \mathbf{d}_r), \mathbf{w}^r_i) \mapsto \bar{\mathbf{w}}^r_i,
\end{equation}
where $\sigma$ is the learnable parameters. In multi-reference frames case, feature vectors from each frame are then aggregated with an average pooling operator as the final image feature vector $\mathbf{w}_i=(\sum_{r=1}^{N}\bar{\mathbf{w}}^r_i)/N$. With image feature as conditional input, the color and semantic representation Eq. \ref{eq:color nerf} and Eq. \ref{eq:semantic nerf} can be rewirtten as:
\begin{equation}
\label{eq:re color nerf}
    f_\phi(\gamma(\mathbf{x}_i), \mathbf{h}_i, \mathbf{w}_i) \mapsto \hat{\mathbf{c}}_i,
\end{equation}
\begin{equation}
\label{eq:re semantic nerf}
    f_\tau(\gamma(\mathbf{x}_i), \mathbf{h}_i, \mathbf{w}_i) \mapsto \hat{\mathbf{s}}_i.
\end{equation}
Color  $\hat{\mathbf{c}}(\mathbf{\mathbf{u}}; \hat{\xi}_t, \theta)$ and semantic logits  $\hat{\mathbf{s}}(\mathbf{\mathbf{u}}; \hat{\xi}_t, \theta)$ are rendered as in Eq.\ \ref{eq:rendering}. The photometric and semantic loss are defined as:
\begin{equation}
\label{eq:color loss}
    \mathcal{L}_{p}=\sum_{\mathbf{u}\in|I|}\left \| \mathbf{c}(\mathbf{u})- \mathbf{\hat{c}}(\mathbf{\mathbf{u}}; \hat{\xi}_t, \theta)\right \|_2,
\end{equation}
\begin{equation}
\label{eq:semantic loss}
    \mathcal{L}_{s}=-\sum_{\mathbf{u}\in|I|}\sum_{k=1}(\mathbf{s}^k(\mathbf{u})\log(\mathbf{\hat{s}}^k(\mathbf{\mathbf{u}}; \hat{\xi}_t, \theta))),
\end{equation}
where $\mathbf{c}(\mathbf{u}),  \mathbf{u}(\mathbf{x})$ are ground-truth color and semantic logits. For the photometric loss $\mathcal{L}_{p}$, we use the $L_2$ norm and for the semantic loss $\mathcal{L}_{s}$, we use multi-class cross-entropy loss. %to encourage the rendered semantic labels to be consistent with provided labels.

\subsection{Self-Supervised Coarse Scene Representation}
\label{subsec:coarse nerf}
SLAM requires real-time performance, but our multi-class scene representations can become hardware-intense on smaller-compute devices. 
To tackle this, we exploit the fact that while a single MLP is not ideal to represent an entire scene due to catastrophic forgetting, it can however be used to represent a small neighborhood of the scene in a lightweight manner and is hence an ideal candidate for tracking. 
%Analyzing the fact that a single MLP has limited capacity to express the entire scene, while tracking only cares about current view, 
%We intentionally let it forget previous knowledge and focus on recent views. 
Thus, we introduce a lightweight coarse scene representation:
\begin{equation}
\label{eq:coarse}
    f_\varphi^c(\gamma(\mathbf{x}_i), \mathcal{V}_\alpha(\mathbf{x}_i)) \mapsto (\mathbf{h}_i^c, \hat{o}_i^c).
\end{equation}
Instead of training this coarse scene representation separately, which is time consuming, latent multi-class scene representations are used to supervise it during mapping:
\begin{equation}
\label{eq:c2f loss}
    \mathcal{L}_{l}=\sum_{i=1}\left \| \mathbf{h}_i- \mathbf{h}_i^c\right \|_2,
\end{equation}
where $\mathbf{h}_i$ is latent vectors from Eq. \ref{eq:fine nerf}. $L_2$ norm is applied on this self-supervised latent loss $\mathcal{L}_{l}$. 

\subsection{Occupancy Probability Approximation}
To achieve accurate and smooth reconstructions with detailed geometry, we propose an occupancy probability approximation loss inspired by the approximate SDF and feature smoothness losses in~\cite{wang2023co}. Specifically, for samples within the truncation region, i.e., points where $|d_i-d(\mathbf{u})|\le tr$, we assume the occupancy probability follows a Gaussian distribution with mean value at the observed distance:
\begin{equation}
    {o}_i=f(|d_i-d(\mathbf{u})|; \mu, \sigma) = \frac{1}{\sigma \sqrt{2\pi}} e^{-\frac{1}{2}\left(\frac{|d_i-d(\mathbf{u})| - \mu}{\sigma}\right)^2}.
\end{equation}
We use this approximation as the ground-truth occupancy value for supervision:
\begin{equation}
    \mathcal{L}_{o} = \sum_{\mathbf{u}\in|I|} \sum_{|d_i-d(\mathbf{u})|\le tr} \left \| {o}_i- \hat{o}_i\right \|_2.
\end{equation}
For points that are far from the surface $|d_i-d(\mathbf{u})| > tr$, we apply a free-space loss~\cite{NiemeyerMOG20} which forces the occupancy to be zero:
\begin{equation}
    \mathcal{L}_{fs} = \sum_{\mathbf{u}\in|I|} \sum_{|d_i-d(\mathbf{u})|> tr} \left \| \hat{o}_i\right \|_2.
\end{equation}

\subsection{Mapping}
\label{subsec:mapping}
Our model consisting of a feature grid $\mathcal{V}_\alpha$ and scene representations $\theta=\{\alpha, \psi^k, \phi, \tau, \varphi\}$ are randomly initialized at the beginning. During the whole process, 2D feature encoder $\mathcal{E}$ is fixed. For the first input frame, camera pose is fixed and only the hash grid and scene representations are optimized. For subsequent frames, scene representations and camera poses are optimized jointly and iteratively every $k$ frames. 

\noindent \textbf{Keyframe Selection and BA:} We follow~\cite{zhu2022nice} to choose the keyframes and frames for bundle adjustment (BA). For each mapping step, bundle adjustment is applied with current frame, the latest keyframe, and $W-2$ selected keyframes. Instead of only carrying local BA, which means selecting $W-2$ keyframes which has overlaps with current frame, we also randomly choose $W-2$ keyframes from the whole keyframes for global BA. 

\noindent \textbf{Reference Frame Selection.}
We denote these selected BA frames as target frames. For each target frame, two reference frames are chosen by following strategy: 
\begin{itemize}
\item for current frame, the two latest keyframes are chosen,
\item for latest keyframe, two previous keyframes are chosen,
\item for other keyframe, one previous keyframe and one later keyframe are chosen.
\end{itemize}

\noindent \textbf{Ray Sampling:}
Fisrt, $R$ pixels are randomly sampled from $W$ target frames. $60\%$ of them are randomly sampled among the image plane, $40\%$ of them are randomly sampled among each class. Then, as describeed in Sec. \ref{subsec:overview}, points are sample along the rays. 
We use depth guided sampling such that $M_s$ points are sampled near the surface and $M_a$ points are sampled in free space.
%Same as in previous methods \cite{zhu2022nice, wang2023co}, with depth guiding, $M_s$ points are sampled near the surface and $M_a$ points are sampled in free space.

\noindent \textbf{New Class Initialization:}
If we encounter a previously unseen class, a new scene representation will be created and initialized. Then, it will be trained separately for 100 iterations before it joins the multi-class scene representations. This allows the system dynamically adding new class, which is more meaningful in practice.

\noindent \textbf{Mapping Loss:}
Mapping process is performed via minimizing the loss functions with respect to the learnable parameters $\theta$ and camera poses $\xi$. 
The final loss $\mathcal{L}_M$ is 
\begin{equation}
\begin{aligned}
   &\mathcal{L}_M(\theta, \xi) = \\
   &\mathcal{L}_{geo}+\lambda_p\mathcal{L}_{p}+\lambda_s\mathcal{L}_{s}+\lambda_l\mathcal{L}_{l} + \lambda_o\mathcal{L}_{o} + \lambda_{fs}\mathcal{L}_{fs}
   \end{aligned}
\end{equation}
where $\lambda_p, \lambda_s, \lambda_l, \lambda_o, \lambda_{fs}$ are respective loss weighting factors.

\subsection{Tracking}
\label{subsec:tracking}
For tracking, the  hash-based feature grid and the scene representations are fixed, and only the camera pose are updated. 
The latest optimized frame is chosen as the reference frame for extracting features and $R_t$ pixels are randomly sampled among the whole image. We use the same ray sampling strategy as in mapping.
Occupancy values $\hat{o}_i^c$ and latent vectors $\mathbf{h}_i^c$ are obtained by Eq.\ \ref{eq:coarse} and latent vectors are passed to Eq.\ \ref{eq:re color nerf} and Eq.\ \ref{eq:re semantic nerf} to obtain the color $\hat{\mathbf{c}}_i$ and semantic values $\hat{\mathbf{s}}_i$.

\noindent \textbf{Tracking Loss:} A modified version of our geometry loss function is used in tracking, where the geometry term is weighted by the standard deviation of the depth prediction $\hat{d}_{var}$:
\begin{equation}
    \mathcal{L}_{geo}^{track}=\sum_{\mathbf{u}\in|I|}\frac{\left \| d(\mathbf{u})- \hat{d}(\mathbf{u}; \hat{\xi}_t, \theta)\right \|_1}{{\sqrt{\hat{d}_{var}(\mathbf{u}; \hat{\xi}_t, \theta)}}},
\end{equation}
where $\hat{d}_{var}(\mathbf{u}; \hat{\xi}_t, \theta)=\sum_{i=1}^M{w_i(\hat{d}(\mathbf{u}; \hat{\xi}_t, \theta)-\hat{d}_i)^2}$.

Given an initial guess of current camera pose and under the constant speed assumption~\cite{zhu2022nice},  camera poses are iteratively updated by minimizing the following loss $\mathcal{L}_T$:
\begin{equation}
    \mathcal{L}_T({\xi}) = \mathcal{L}_{geo}^{track}+\lambda_p\mathcal{L}_{p}+\lambda_s\mathcal{L}_{s},
\end{equation}
where $\xi$ is the camera poses and $\lambda_p, \lambda_s$ are respective loss weighting factors.

\section{Experiments}
\subsection{Experimental Setup}
\noindent \textbf{Datasets}
We evaluate our method on both synthetic and real-world datasets with semantic maps. Following other neural implicit SLAM methods, for the reconstruction quality, we evaluate quantitatively on 8 synthetic scenes from Replica \cite{straub2019replica} and qualitatively on 6 scenes from ScanNet \cite{dai2017scannet}. As for camera pose accuracy, we evaluate quantitatively on both Replica and ScanNet. The ground-truth poses of Replica are from simulation, while ground-truth poses of ScanNet are obtained with BundleFusion \cite{dai2017bundlefusion}.

\noindent \textbf{Metrics}
We evaluate the reconstruction quality using $Depth L1(cm), Accuracy(cm), Completion(cm)$ and $Completion Ratio(\%)$ with a threshold of 5cm. For evaluation of camera tracking, we adopt $ATE\ RMSE(cm)$ \cite{sturm2012benchmark}. To evaluate our semantic reconstruction, we report $mIoU$.

\noindent \textbf{Baselines}
We compare against the state-of-the-art methods NICE-SLAM \cite{zhu2022nice}, Co-SLAM \cite{wang2023co}, ESLAM \cite{johari2023eslam}, Point-SLAM \cite{sandstrom2023point} and ADFP \cite{Hu2023LNI-ADFP} as our main baselines for reconstruction quality and camera tracking. For semantic evaluation, we compare against the very recent NIDS-SLAM \cite{haghighi2023neural} as all other methods do not predict semantic information.

\noindent \textbf{Implementation Details}
%We run our method on multiple GPUs for multiple runs, including NVIDIA A5000, 2080ti and Titan Xp. 
We perform single GPU training (NVIDIA 2080ti), and for experiments with default settings, we use $R_t=500$ pixels with 30 iterations for tracking and $R=2000$ pixels for mapping with 100 iterations on Replica, and 200 iterations on ScanNet. For our multi-resolution hash grid settings, we follow Co-SLAM \cite{wang2023co}. We use two-layer MLPs with 32 hidden units for each scene representation. We use a learning rate of 0.005 and 0.001 for all learnable parameters on Replica and ScanNet, respectively. For camera poses, we use a learning rate of 0.001 in tracking and 0.0005 in mapping.  We use the first layer of ResNet \cite{he2016deep} output as our 2D features. Please see the sup.\ mat.\ for additional details.

\begin{table*}
    \centering
    \fontsize{6pt}{8pt}\selectfont
    \begin{tabular}{@{}l |c @{\hspace{0.4cm}}c @{\hspace{0.4cm}}c @{\hspace{0.4cm}}c @{\hspace{0.4cm}}c @{\hspace{0.4cm}}c |c |c @{\hspace{0.4cm}}c@{\hspace{0.4cm}}c@{\hspace{0.4cm}}c@{\hspace{0.4cm}}c@{\hspace{0.4cm}}c@{\hspace{0.4cm}}c@{\hspace{0.4cm}}c |c}
    \toprule
          & \multicolumn{7}{c|}{ScanNet \cite{dai2017scannet}} & \multicolumn{9}{c}{Replica \cite{straub2019replica}}\\
          & 0000 & 0059 & 0106 & 0169 & 0181 & 0207 & Avg. & room\_0 & room\_1 & room\_2 & office\_0 & office\_1 & office\_2 & office\_3 & office\_4 & Avg.\\
    \midrule
         NICE-SLAM \cite{zhu2022nice} & 8.64 & 12.25 & \cellcolor[RGB]{255, 251, 198}8.09 & 10.28 & \cellcolor[RGB]{255, 251, 198}12.93 & \cellcolor[RGB]{233, 241, 188}5.59 & 9.63 
                  & 1.69 & 2.04 & 1.55 & 0.99 & 1.39 & 3.97 & 3.08 & 1.95 & 2.08\\
         Co-SLAM \cite{wang2023co} & \cellcolor[RGB]{233, 241, 188}7.18 & 12.29 & 9.57 & \cellcolor[RGB]{233, 241, 188}6.62 & 13.43 & 7.13 & 9.37
                 & \cellcolor[RGB]{255, 251, 198}0.65 & 1.13 & 1.43 & 0.55 & 0.50 & \cellcolor[RGB]{233, 241, 188}\textbf{0.46} & 1.40 & 0.77 & 0.86\\
         ESLAM \cite{johari2023eslam}  & \cellcolor[RGB]{255, 251, 198}7.3 & \cellcolor[RGB]{255, 251, 198}8.5 & \cellcolor[RGB]{233, 241, 188}7.5 & \cellcolor[RGB]{202, 229, 205}\textbf{6.5} & \cellcolor[RGB]{202, 229, 205}\textbf{9.0} & 5.7 & \cellcolor[RGB]{233, 241, 188}7.42
               & 0.76 & \cellcolor[RGB]{255, 251, 198}0.71 & \cellcolor[RGB]{255, 251, 198}0.56 & \cellcolor[RGB]{255, 251, 198}0.53 & \cellcolor[RGB]{255, 251, 198}0.49 & 0.58 & \cellcolor[RGB]{255, 251, 198}0.74 & \cellcolor[RGB]{255, 251, 198}0.64 & \cellcolor[RGB]{255, 251, 198}0.62\\
         Point-SLAM \cite{sandstrom2023point} & 10.24 & \cellcolor[RGB]{233, 241, 188}7.81 & 8.65 & 22.16 & 14.77 & 9.54 & 13.97
                     & \cellcolor[RGB]{233, 241, 188}0.61 & \cellcolor[RGB]{202, 229, 205}\textbf{0.41} & \cellcolor[RGB]{202, 229, 205}\textbf{0.37} & \cellcolor[RGB]{233, 241, 188}0.38 & \cellcolor[RGB]{233, 241, 188}0.48 & \cellcolor[RGB]{255, 251, 198}0.54 & \cellcolor[RGB]{233, 241, 188}0.72 & \cellcolor[RGB]{233, 241, 188}0.63 & \cellcolor[RGB]{233, 241, 188}0.52\\
         ADFP \cite{Hu2023LNI-ADFP} & - & 10.50 & \cellcolor[RGB]{202, 229, 205}\textbf{7.48} & 9.31 & - & \cellcolor[RGB]{255, 251, 198}5.67 & \cellcolor[RGB]{255, 251, 198}8.24
                & 1.39 & 1.55 & 2.60 & 1.09 & 1.23 & 1.61 & 3.61 & 1.42 & 1.81\\
         \textbf{Ours} & \cellcolor[RGB]{202, 229, 205}\textbf{5.42} & \cellcolor[RGB]{202, 229, 205}\textbf{5.20} & 9.11 & \cellcolor[RGB]{255, 251, 198}7.70 &
                        \cellcolor[RGB]{233, 241, 188}10.12 & \cellcolor[RGB]{202, 229, 205}\textbf{4.91} & \cellcolor[RGB]{202, 229, 205}\textbf{7.07} 
                      & \cellcolor[RGB]{202, 229, 205}\textbf{0.49} & \cellcolor[RGB]{233, 241, 188}0.46 & \cellcolor[RGB]{233, 241, 188}0.38 & \cellcolor[RGB]{202, 229, 205}\textbf{0.34} & \cellcolor[RGB]{202, 229, 205}\textbf{0.35} & \cellcolor[RGB]{202, 229, 205}\textbf{0.39} & \cellcolor[RGB]{202, 229, 205}\textbf{0.62} & \cellcolor[RGB]{202, 229, 205}\textbf{0.60} & \cellcolor[RGB]{202, 229, 205}\textbf{0.45}\\
    \bottomrule
    \end{tabular}
    \caption{
    \textbf{Quantitative Tracking Comparison.}
    We show ATE RMSE$\downarrow$ (cm)  results of an average of 3 runs on ScanNet~\cite{dai2017scannet} and Replica \cite{straub2019replica}. We find that our method leads to state-of-the-art performance, improving tracking results on both synthetic and real-world data.
    }
    \label{tab:ate}
\end{table*}

\subsection{Tracking Evaluation}
\begin{figure}
\includegraphics[width=8.3cm]{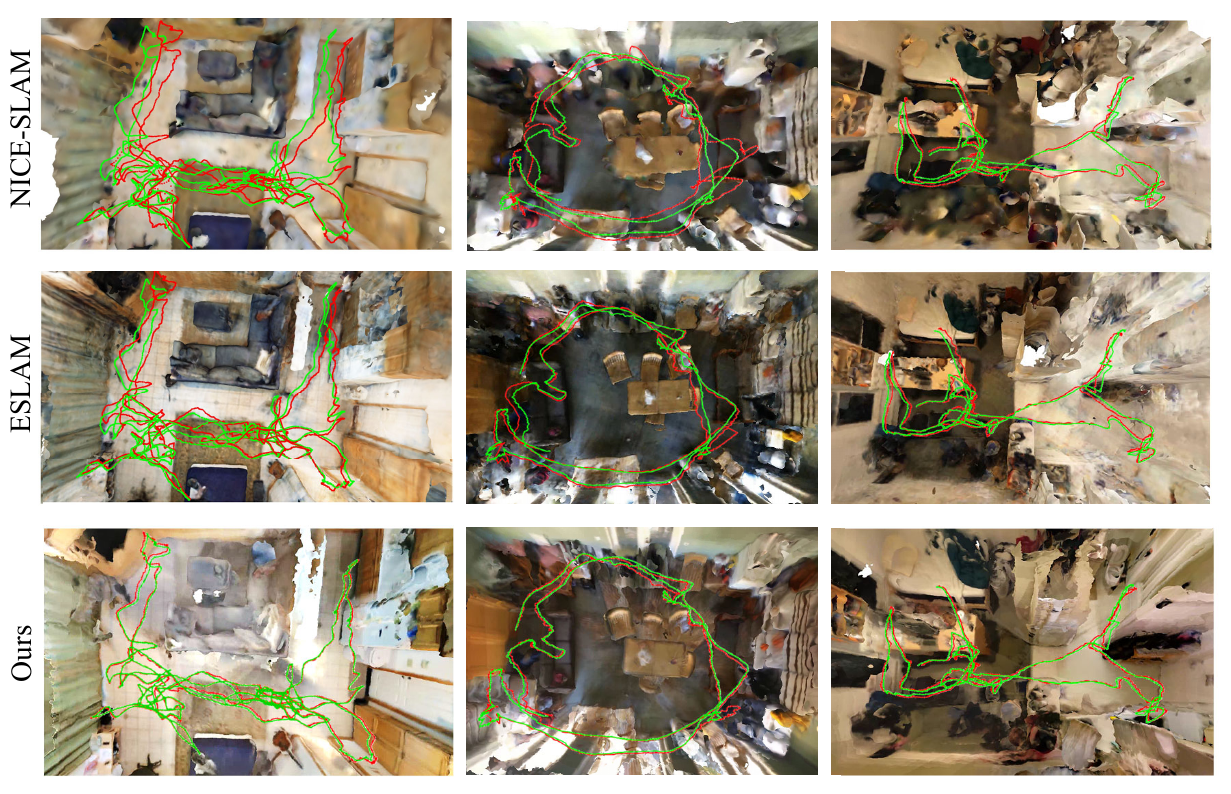}
\centering
\caption{\textbf{Qualitative Comparison on ScanNet.} The ground-truth camera trajectory is shown in \textcolor{green}{green} and the estimated trajectory is shown in \textcolor{red}{red}. Our method predicts more accurate camera trajectories and does not suffer from pose drifting.}
\label{fig:scannet vis}
\end{figure}

We show quantitative results on both Replica \cite{straub2019replica} and ScanNet \cite{dai2017scannet} in Table \ref{tab:ate}. Our method demonstrates state-of-the-art tracking performance on the Replica dataset, outperforming other methods over 10\% on average in terms of ATE RMSE. It is important to highlight that while both our method and Point-SLAM \cite{sandstrom2023point} achieve impressive tracking results on Replica, our method proves to be more robust and effective, particularly in complex real-world scenarios like those encountered in the ScanNet dataset. We attribute this is to the fact that Point-SLAM strongly relies on accurate depth supervision, while our approximated occupancy probability formulation provides reliable supervision and our multi-view consistency constraints can effectively address depth map ambiguities even if the depth input is not accurate. Fig.\ \ref{fig:scannet vis} shows a qualitative comparison on ScanNet. Compared with NICE-SLAM \cite{zhu2022nice} and ESLAM \cite{johari2023eslam}, our method suffers from less trajectory drifting and, especially for those scenes with more texture, our method can improve ATE RMSE significantly. This highlights the adaptability and robustness of our approach, especially in challenging scenarios, making it a promising solution for practical applications. 

\subsection{Reconstruction Evaluation}
\begin{figure*}
\includegraphics[width=17cm]{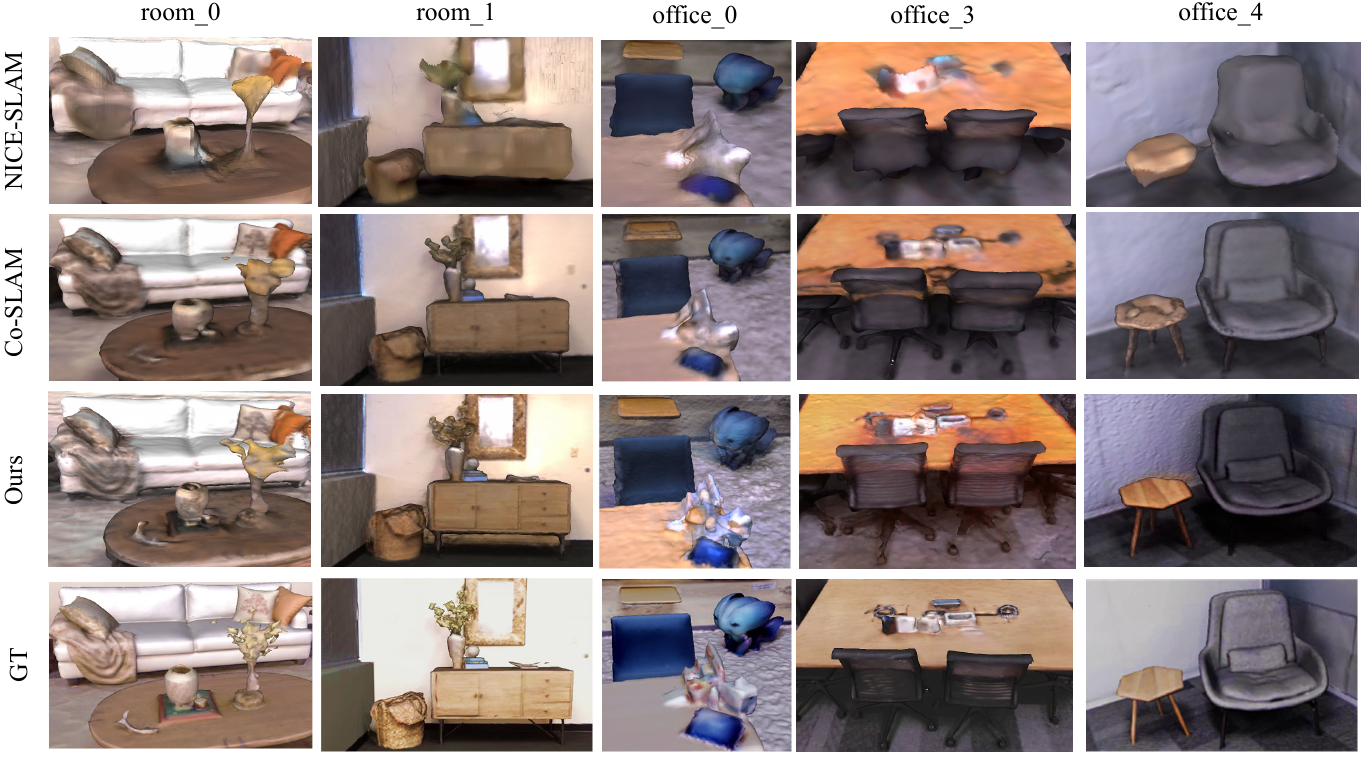}
\centering
\caption{\textbf{Qualitative Reconstruction on Replica.} Our method achieves reconstructions with more geometric and appearance details.}
\label{fig:replica_rec}
\end{figure*}

Qualitative and quantitative reconstruction results on Replica \cite{straub2019replica} are shown in Fig.\ \ref{fig:replica_rec} and Table\ \ref{tab:rec replica}, respectively. 
%Co-SLAM \cite{wang2023co} is a SDF-based method, which represents shapes and surfaces in a compact and continuous manner. 
%Thus, they tend to generate continues but smooth method. Table \ref{tab:rec replica} shows that even though Co-SLAM achieves high score in quantitative comparison, but for the qualitative comparison, we find that Co-SLAM lack of many texture as NICE-SLAM. 
We find that SDF-based Co-SLAM~\cite{wang2023co} leads to good but overly-smoothed reconstructions and both Co-SLAM and NICE-SLAM~\cite{zhu2022nice} do not lead to satisfactory texture reconstruction.
%Our method and NICE-SLAM \cite{zhu2022nice} are occupancy-based method, which often used for more discrete representations of the environment, especially in the context of SLAM. 
%However, occupancy's binary nature can introduce noise, particularly in regions near the surface or occlusions where the decision of occupancy is ambiguous.
%Therefore, while our method doesn't stand out in terms of metrics, we can still provide reconstruction with much more details compared with Co-SLAM, even for very tiny objects, which are disappeared in other methods.
In contrast, our method is able to reconstruct fine geometric structures as well as appearance details (see e.g.\ the flowers on the table of "room\_1" or the side table of "office\_4" in Fig.\ \ref{fig:replica_rec}). 
While we improve also quantitatively over NICE-SLAM, we find that the reconstruction metrics favor the overly-smoothed reconstructions from the SDF-based methods such as Co-SLAM, mainly due to the fact that errors in unobserved regions (such as under a table) dominate the evaluation metrics. 
We plan to investigate appropriate regularization techniques for unobserved regions, e.g.\ based on diffusion priors, for our image-based approach in the future. 
%The quantitative comparison of reconstruction accuracy on Replica is reported in Table \ref{tab:rec replica} in average of 8 scenes. 

\begin{figure}
\includegraphics[width=8.2cm]{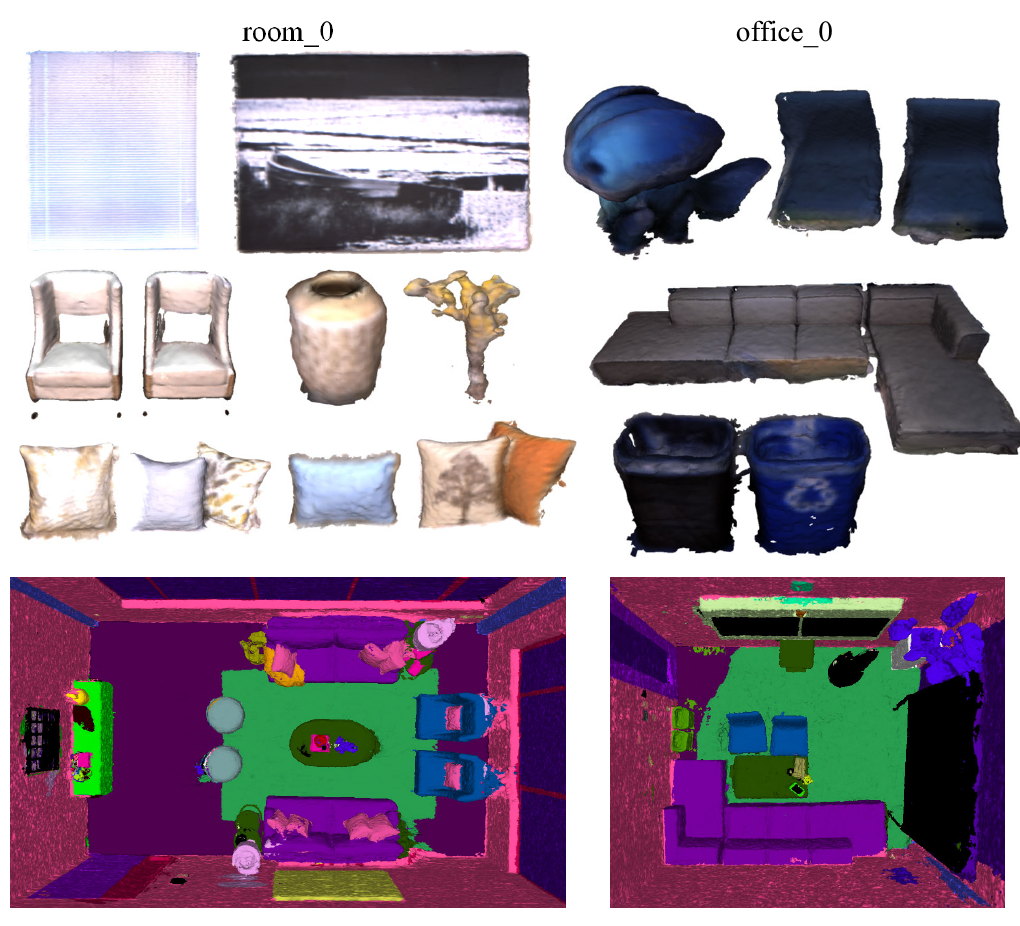}
\centering
\caption{\textbf{Decomposed Reconstruction on Replica.} We show the 3D semantic mesh (bottom) and its decomposition with RGB colors (top) of two scenes from Replica.}
\label{fig:replica semantic}
\end{figure}

\begin{table}
    \centering
    \fontsize{6pt}{7pt}\selectfont
    \begin{tabular}{@{}l |c |c  @{\hspace{0.3cm}}c @{\hspace{0.3cm}}c @{\hspace{0.3cm}}c}
    \toprule
          & Method & Depth L1$\downarrow$ & Acc.$\downarrow$ & Comp.$\downarrow$ & Comp. Ratio$\uparrow$ \\
    \midrule
         \multirow{2}{*}{SDF-Based} & Co-SLAM & 1.58 & 2.10 & 2.08 & 92.99 \\
                                    & ESLAM  & \textbf{1.18} & \textbf{0.97} & \textbf{1.05} & \textbf{98.60} \\
    \midrule
         \multirow{3}{*}{Occupancy-Based} & NICE-SLAM & 3.53 & 2.85 & 3.00 & 89.33 \\
                                          & ADFP & 1.81 & 2.59 & 2.28 & 93.38\\
                                          & \textbf{Ours} & 3.16 & 2.76 & 2.74 & 91.73 \\
    \bottomrule
    \end{tabular}
    \caption{\textbf{Quantitative Reconstruction Comparison on Replica.}
    %of an average of 3 runs on Replica.
    }
    \label{tab:rec replica}
\end{table}

\subsection{Semantics Evaluation}
%To evaluate the semantic segmentation accuracy, we render a semantic map at each view.
Table \ref{tab:semantic} shows the quantitative evaluation of our method in comparison to another neural semantic SLAM method NIDS-SLAM \cite{haghighi2023neural}. We follow NIDS-SLAM and report the mIoU on four scenes.
Our method achieves better results despite NIDS-SLAM using ORB-SLAM \cite{mur2015orb} as its tracking processor, and we are the first neural SLAM method to achieve simultaneous localization, reconstruction and segmentation all at once. Fig.\ \ref{fig:replica semantic} shows class-wise reconstruction results on Replica dataset. Our method can represent and reconstruct each class separately leading to decomposed representations, which can be used as a 3D prior for further downstream tasks.
\begin{table}
    \centering
    \fontsize{6pt}{7pt}\selectfont
    \begin{tabular}{@{}l |c @{\hspace{0.3cm}}c @{\hspace{0.3cm}}c @{\hspace{0.3cm}}c @{\hspace{0.3cm}}c}
    \toprule
          Scene & room\_0 & room\_1 & room\_2 & office\_0 & Avg. \\
    \midrule
         NIDS-SLAM & 82.45 & 84.08 & 76.99 & \textbf{85.94} & 82.37 \\
         \textbf{Ours} & \textbf{88.32} & \textbf{84.90} & \textbf{81.20} & 84.66 &  \textbf{84.77}\\
    \bottomrule
    \end{tabular}
    \caption{
    \textbf{Semantic Segmentation Comparison on Replica.}
    %of an average of 3 runs on Replica.}
    }
    \label{tab:semantic}
\end{table}

\subsection{Runtime Analysis}
We report runtimes on Replica's room\_0 scene in Table \ref{tab:runtime}, using a NVIDIA 2080ti GPU. The tracking and mapping time are reported per frame. We observe that our approach effectively balances performance and runtime considerations, enabling high-fidelity as well as fast SLAM.

\begin{table}
    \centering
    \fontsize{6pt}{7pt}\selectfont
    \begin{tabular}{@{}l |c @{\hspace{8pt}}c @{\hspace{8pt}}c @{\hspace{8pt}}c @{\hspace{8pt}}c @{\hspace{8pt}}c}
    \toprule
         Per frame & NICE-SLAM & Co-SLAM & Point-SLAM & \textbf{Ours}\\
    \midrule
         Tracking & 1.32 s & 0.12 s & 0.85 s & 0.36 s\\
         Mapping & 10.92 s & 0.33 s & 9.85 s & 7.58 s \\
    \bottomrule
    \end{tabular}
    \caption{
    \textbf{Runtime Comparison on Replica.}
    %of an average of 3 runs on Replica.}
    }
    \label{tab:runtime}
\end{table}

\begin{table}
    \centering
    \fontsize{6pt}{7pt}\selectfont
    \begin{tabular}{@{}l |c @{\hspace{8pt}}c @{\hspace{8pt}}c @{\hspace{8pt}}c @{\hspace{8pt}}c}
    \toprule
         Method & w/o multi-class & w/o 2D feature & w/o init. & w/o occ. approx. & full model\\
    \midrule
         Replica & 1.16 & 0.71 & 0.53 & 0.86 & 0.49\\
         ScanNet  & 7.52 & 6.22 & 174.4 & 5.77 &  5.20\\
    \bottomrule
    \end{tabular}
    \caption{
    \textbf{Quantitative Ablation Study (ATE RMSE).}
    %Ablation study results on ATE RMSE.
    }
    \label{tab:ablation}
\end{table}

\subsection{Ablation Study}
In Table \ref{tab:ablation} we report quantitative results for our method where we ablate various components which we discuss in greater detail in the following. We test on scene room\_0 from Replica and scene 0059 from ScanNet.

\noindent \textbf{New Class Initialization.}
We find that if a new class representation is not initialized appropriately before it is added to the system, it leads to inadequate RGB predictions not only for the new class but also for existing classes due to a bleeding effect.
%not only pred also perturb predictions for existing classes. %Table \ref{tab:ablation} shows the quantitatively evaluation on room 0 from Replica and scene0059 from ScanNet. 
It is important to note that this in turn also leads to an increased camera pose error 
%also increases when a new class is added, and can then influence the following steps, 
and sometimes causing further tracking failures. This is in particular visible on the real-world dataset ScanNet.

\noindent \textbf{Multi-Class Scene Representations.}
We ablate our multi-class scene representations 
%on room 0 from Replica and scene0059 from ScanNet 
by using only a single representation during the whole process. Table \ref{tab:ablation} shows that our full mode leads to higher accuracy considering ATE RMSE than only using a single geometry representation. The difference is larger on ScanNet where the input data is noisy and the size of the scene is larger compared to Replica.

\begin{figure}
\includegraphics[width=8.5cm]{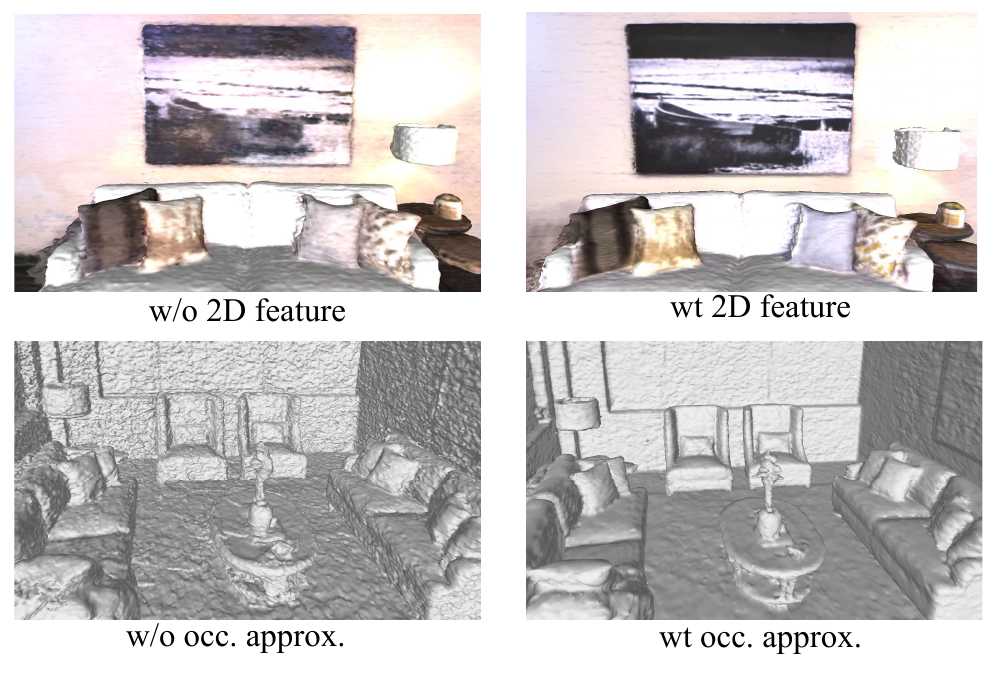}
\centering
\caption{\textbf{Ablation Study.} We show that 2D image feature can provide conditional information for reconstruction (top). And occupancy probability approximation can help to smooth the reconstruction (bottom).}
\label{fig:ab_occ}
\end{figure}

\noindent \textbf{Conditional Image Feature Input.}
Image features serve as a conditional input for our system, enforcing a multi-view geometry constraint and enhancing the texture prediction. Table \ref{tab:ablation} shows a decline in ATE RMSE on both synthetic and real-world datasets when removing the conditional input. Fig.\ \ref{fig:ab_occ} shows a visual comparison of the reconstructed meshes. We find that without 2D image features, the reconstruction become smoother and contain less details.

\noindent \textbf{Occupancy Probability Approximation.}
Finally, We investigate the effectiveness of our occupancy probability approximation and show a qualitative comparison in Fig.\ \ref{fig:ab_occ}. The output mesh tends to be noisier and artifacts appear.
In contrast, the meshes of our full model are smoother while containing fine details.
%, while 
%the depth to approximate occupancy value for every point in the space can serve as a smooth supervision. 
Table \ref{tab:ablation} further shows that our full model also leads to better tracking performance.
%on tracking for synthetic dataset where the depth are absolute accurate, while it does not help as much on real-world datasets.

%% file: 10_conclusion.tex
\section{Conclusion}
\label{sec:conclusion}
We present DNS SLAM, the first method that uses a neural representation for simultaneously localization, reconstruction and segmentation.
We propose to use a semantically decomposed scene representation in combination with conditional 2D image features, enforcing stronger geometric constraints and hence enabling better tracking and more detailed appearance prediction while being more robust to noisy depth inputs.
%We show that using multi-class scene representation and conditional 2D image features can help solve the ambiguity of depth input in real-world dataset and provide more texture for reconstruction. 
Our occupancy probability approximation serves as a strong supervision signal, enabling accurate and smooth mesh reconstruction. Extensive experiments show that our method significantly improves the tracking accuracy compared to state-of-the-art baselines while also providing class-wise decompositions.

\noindent \textbf{Limitations.}
Our method relies on RGB-D and semantic maps as input. While being more robust to noisy depth inputs than previous works, we plan to investigate RGB only input in the future. 
%and use occupancy as output, which can lead to noisy mesh reconstruction. In addition, our method does not explore the class-wise reconstruction for further utilization. They can also serve as landmarks for better tracking.

%\section*{Acknowledgements}
%This work is partially supported by th ... The authors thank the ...

%% file: 12_appendix.tex
\section{Implementation details}
\subsection{Hyperparameters}
In the following, we report implementation details and hyperparameters used for our method to achieve high-accuracy tracking and mapping.

\noindent \textbf{Default Settings.}
 We use $L=16$ level hash grids with $R_{min}=16$ to $R_{max}$, where we set the maximal voxel size to 2cm for determining $ R_{max}$. 
 We use 16 bins for the OneBlob encoding of each dimension.
For the network heads, we use two-layer MLPs with 32 hidden units for each scene presentation. The loss weights we set to $\lambda_p=3,\ \lambda_s=0.1,\ \lambda_l=10,\ \lambda_o=10,\ \lambda_{fs}=5$ and the truncation distance $tr$ is set to 10cm. For ray sampling, we use $M_s=15$ and $M_a=32$ points.
In a forward pass, We sample $R_t=500$ pixels during tracking and $R=2000$ pixels during mapping. In tracking, we estimate camera pose for each frame. In mapping, bundle adjustment will be applied every 5 frame, which updates the scene representations and camera poses jointly. The window size $W$ for bundle adjustment is set to 5 keyframes. For initialization, we start with 500 iterations for the first frame.

\noindent \textbf{Replica.}
For the Replica dataset \cite{straub2019replica}, we perform 30 iterations for tracking and 100 iterations for mapping. We choose a keyframe every 30 frames. We set the learning rate for all learnable parameters to 0.005 except for the camera parameters for which we use a learning rate of 0.001 in tracking and 0.0005 in mapping.

\noindent \textbf{ScanNet.}
For the ScanNet dataset \cite{dai2017scannet}, we use $L=20$ level hash grids with $R_{min}=16$ to $R_{max}$, where we use a maximal voxel size of 4cm to determine $ R_{max}$. We perform 30 iterations of tracking and 200 iterations of mapping iteratively. We choose a keyframe every 15 frames. We set the learning rate for all learnable parameters to 0.001 except for the camera parameters for which we use a learning rate of 0.001 during tracking and 0.0005 during mapping.

\section{Additional Experimental Results}
\subsection{More Reconstruction Results}
\begin{figure*}
\includegraphics[width=17cm]{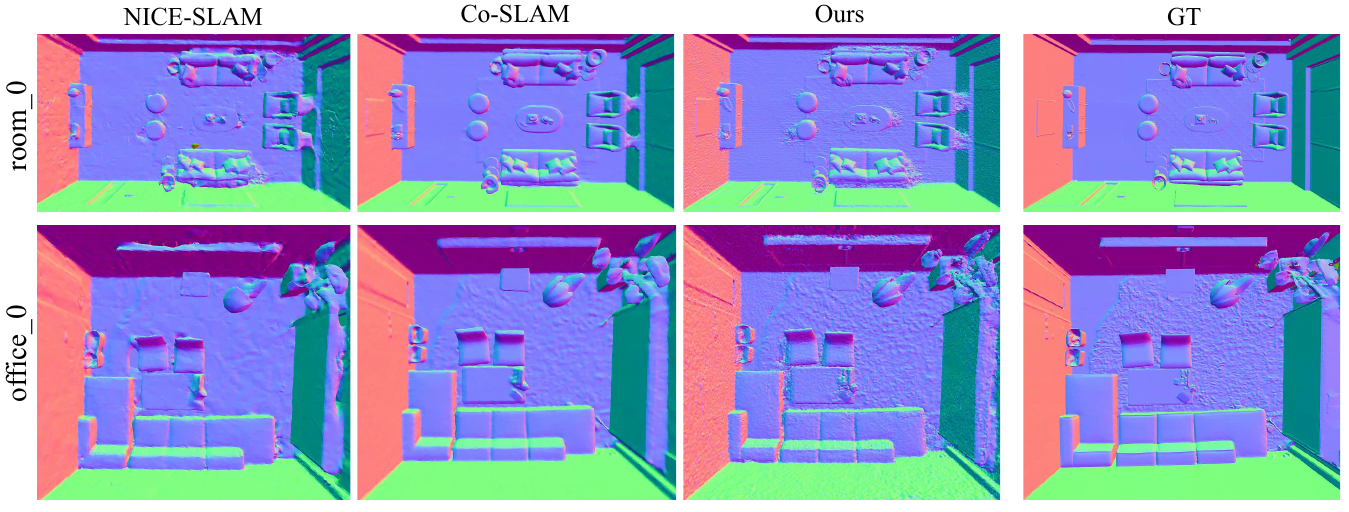}
\centering
\caption{\textbf{Normal Map of Reconstructions on Replica.}}
\label{fig:supp_normal}
\end{figure*}

\begin{figure*}
\includegraphics[width=17cm]{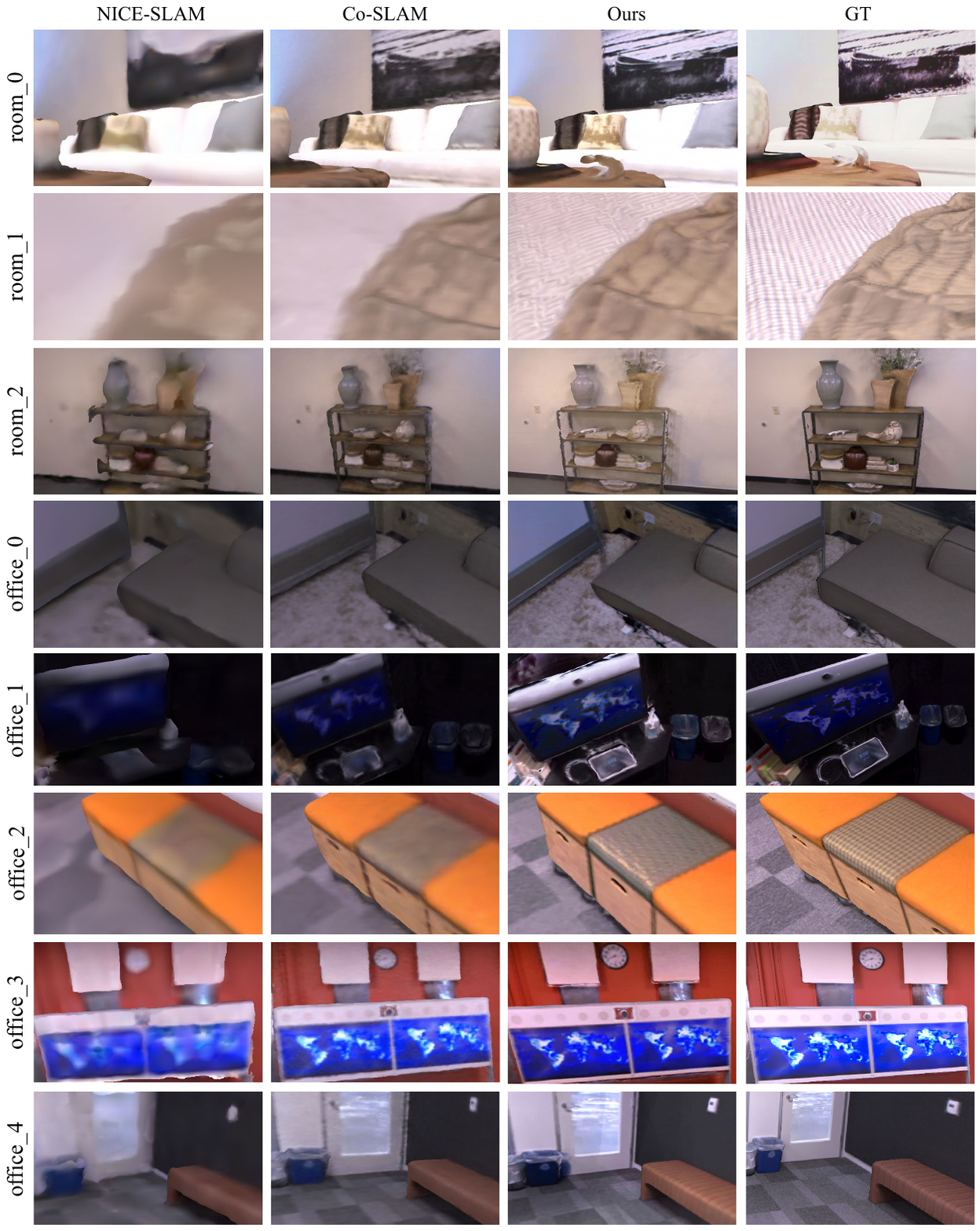}
\centering
\caption{\textbf{Zoom-In View of Reconstructions on Replica.}}
\label{fig:supp_rec_replica}
\end{figure*}

\begin{figure*}
\includegraphics[width=17cm]{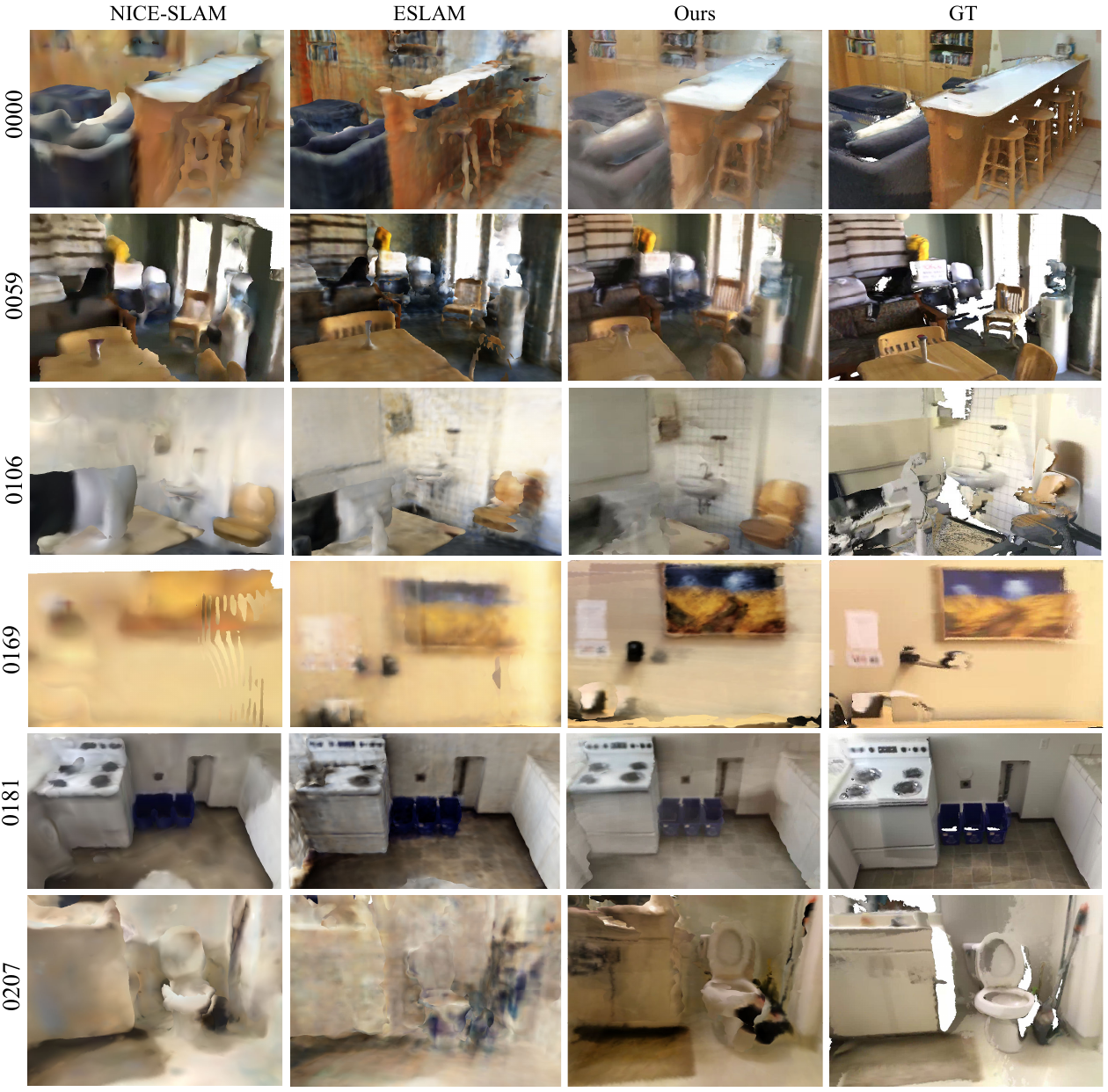}
\centering
\caption{\textbf{Zoom-In View of Reconstructions on ScanNet.}}
\label{fig:supp_rec_scannet}
\end{figure*}

In this section, we report additional reconstruction results. Fig.\ \ref{fig:supp_normal} shows a top-down view comparison and Fig.\ \ref{fig:supp_rec_replica} shows zoom-in views on Replica. Even though SDF-based methods such as Co-SLAM \cite{wang2023co} and ESLAM \cite{johari2023eslam} lead to better results in terms of quantitative metrics, we observe that our method leads to meshes with more geometric details. For example, our method reconstructs the individual objects in the scene (see e.g.\ the tiny object on the table of "room\_0" in Fig.\ \ref{fig:supp_rec_replica}) while the other methods struggle to do so. We also show  reconstruction results on ScanNet. Fig.\ \ref{fig:supp_rec_scannet} shows that our method leadst to better reconstructions while NICE-SLAM \cite{zhu2022nice} and ESLAM \cite{johari2023eslam} show limited reconstruction capabilities on these challenging real-world scenes (see e.g.\ the water dispenser of "0059" and the toilet of "0207" in Fig.\ \ref{fig:supp_rec_scannet}).

\subsection{More Semantic Results on Synthetic Scenes}
\begin{figure*}
\includegraphics[width=13cm]{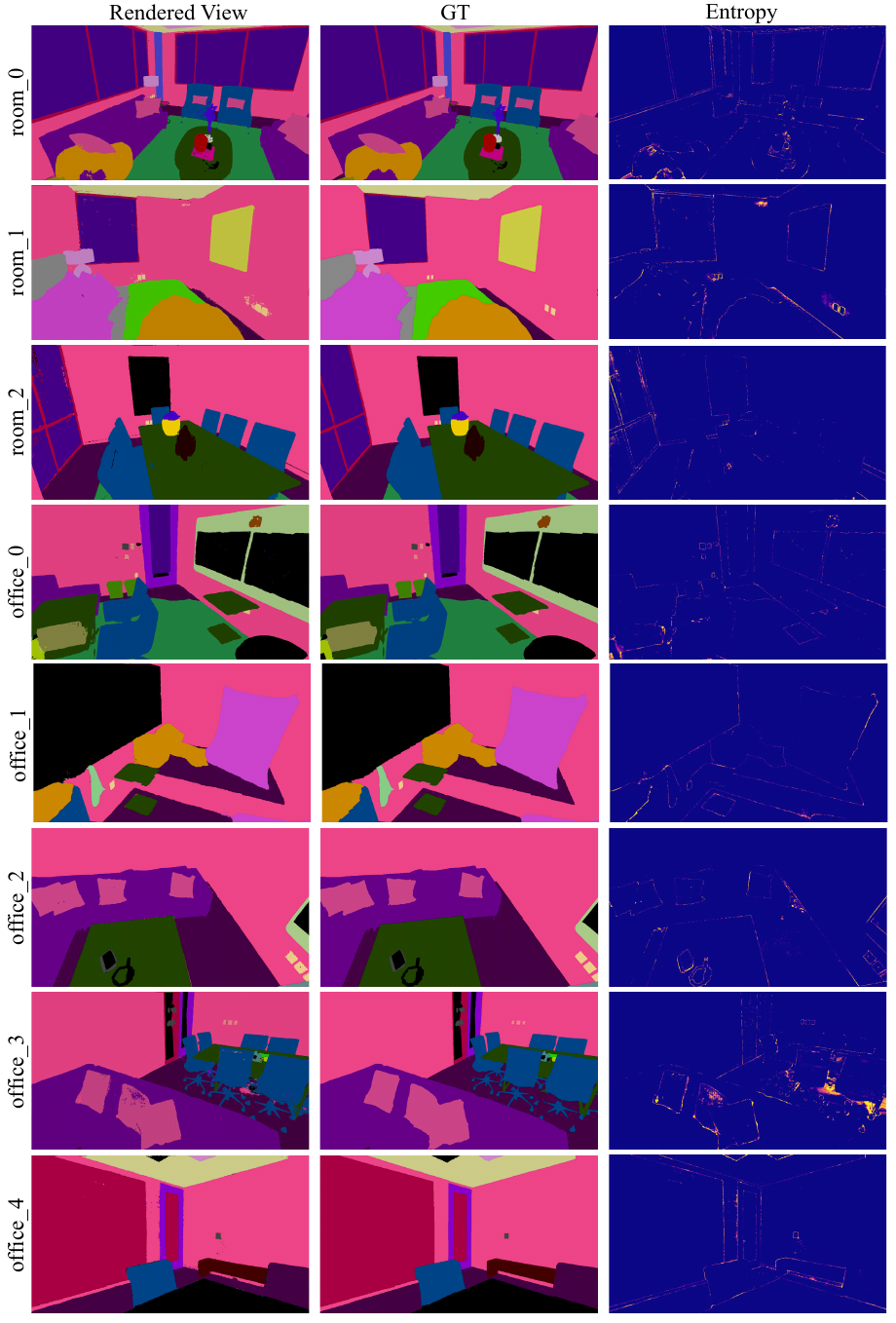}
\centering
\caption{\textbf{Semantic Segmentation on Replica.} From left to right: rendered semantic labels, ground-truth semantic labels and information entropy, highlighting regions with niosy predictions.}
\label{fig:supp_semantic}
\end{figure*}

\begin{figure*}
\includegraphics[width=17cm]{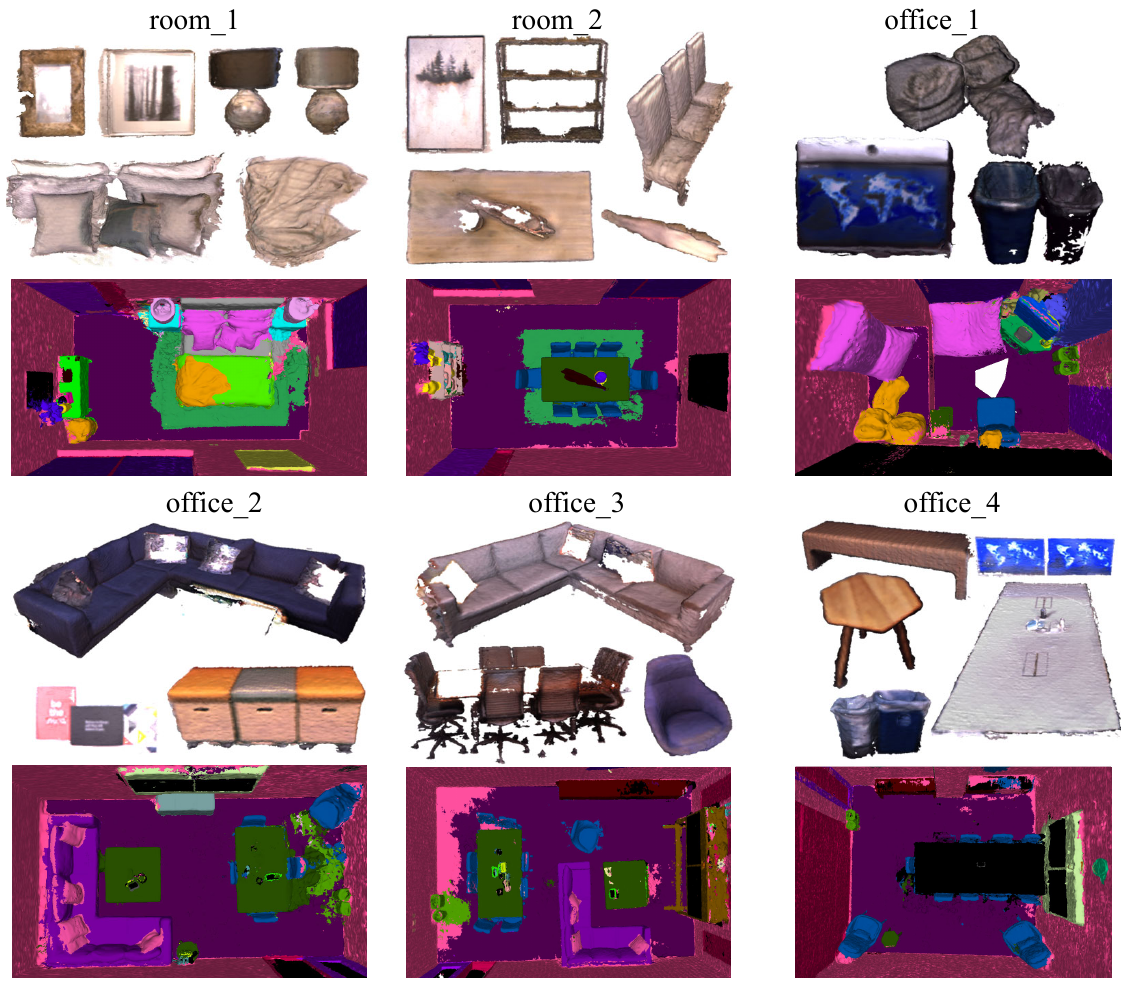}
\centering
\caption{\textbf{Decomposed Reconstruction on Replica}}
\label{fig:supp_decomp}
\end{figure*}

We first show qualitative results for semantic segmentation compared with the ground-truth in Fig.\ \ref{fig:supp_semantic}. We are able to accurately segment the 3D space with errors only occur near the boundaries of objects. Fig.\ \ref{fig:supp_decomp} shows decomposition results of the remaining scenes. Our method enables class-wise decomposed reconstructions with fine geometric details.
%Our method can decompose each class in the scene and reconstruct them with  details even for tiny objects.

\begin{figure}
\includegraphics[width=8cm]{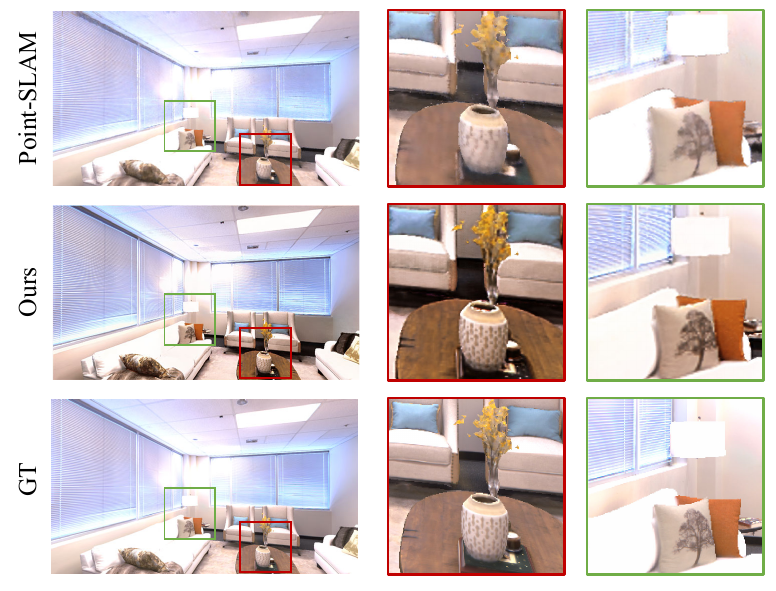}
\centering
\caption{\textbf{Rendering Results on Replica.}}
\label{fig:supp_render}
\end{figure}

\subsection{Novel-View Synthesis Evaluation}
Table \ref{tab:render} shows a comparison for novel view synthesis performance across 8 senes from Replica. We follow the evaluation protocol in~\cite{sandstrom2023point}. Note that our method can reach on par results with Point-SLAM on SSIM and LPIPs \cite{zhang2018unreasonable} without generating intermediate point cloud representations which requires significant computational resource and time. Further, Fig.\ \ref{fig:supp_render} shows that our method better maintains the image structure of the ground truth next to the color, which is especially important for real-world scenes where the input color images can suffer from bad light conditions or motion blur. We hypothesize that this is largely coming from incorporating a pre-trained ResNet \cite{he2016deep} model, which was trained for image recognition, and in turn tends to preserve texture details of images instead of only the RGB color. We plan to implement an additional exposure compensation mechanism to further improve color prediction in future research.

\subsection{Limitation Discussion}
\begin{figure*}
\includegraphics[width=10cm]{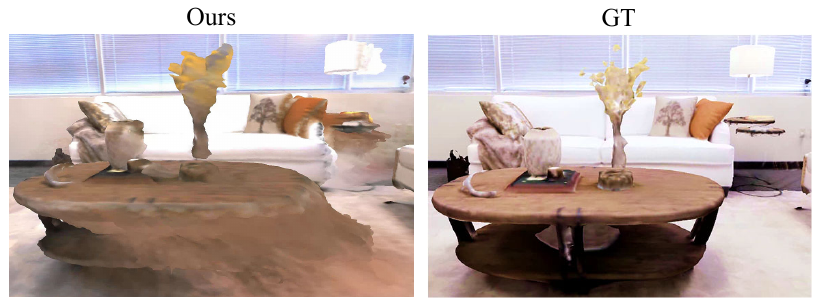}
\centering
\caption{\textbf{Failure Case on Replica.} .}
\label{fig:supp_faliure}
\end{figure*}

Fig.\ \ref{fig:supp_faliure} shows our reconstruction of an unobserved region.
We observe that our occupancy-based approach tends to predict a surface continuation in the unobserved region, while being freespace in the ground truth. We plan to investigate more complex regularization strategies and partial infilling approaches to make our approach more robust for sparser captures.
%We believe this is because that occupancy-based methods only supervise the occupancy probability value of points before the surface. But for points behind the surface, occupancy-based methods tend to predict as being occupied while SDF-based methods tend to be predict as free space.

\begin{table}
    \centering
    \fontsize{7pt}{8pt}\selectfont
    \begin{tabular}{@{}l |c @{\hspace{8pt}}c @{\hspace{8pt}}c @{\hspace{8pt}}c}
    \toprule
         Method & PSNR(dB)$\uparrow$ & SSIM$\uparrow$ & LPIPS$\downarrow$ \\
    \midrule
         NICE-SLAM & 24.42 & 0.809 & 0.233 \\
         Point-SLAM  & \textbf{35.17} & \textbf{0.975} & 0.124 \\
         \textit{Our} & 22.96 & 0.963 & \textbf{0.119} \\
    \bottomrule
    \end{tabular}
    \caption{Rendering Results of an average of 3 runs on Replica.}
    \label{tab:render}
\end{table}